\pdfoutput=1

\RequirePackage{fix-cm}
\documentclass[smallcondensed]{svjour3}     % onecolumn (ditto)
\smartqed  % flush right qed marks, e.g. at end of proof
\usepackage{graphicx}
\usepackage{adjustbox}
\usepackage{graphicx}
\usepackage{hyperref}
\usepackage{natbib}
\usepackage{natbib}
\setcitestyle{square,numbers}
\usepackage[ruled,vlined]{algorithm2e}

\setcitestyle{square}
\begin{document}
\title{Towards a Universal Gating Network for Mixtures of Experts}



\author{Chen Wen Kang       \and
        Chua Meng Hong      \and
        Tomas Maul
}


\institute{Chen Wen Kang \at
              University of Nottingham Malaysia \\
              \email{cwk1998@hotmail.com}           %  \\
          \and
          Chua Meng Hong \at
              University of Nottingham Malaysia \\
             \email{hcycm1@nottingham.edu.my}
        \and
        Tomas Maul \at
              University of Nottingham Malaysia \\
              Tel.: +60 (0)3 8924 8232\\
              \email{Tomas.Maul@nottingham.edu.my}
}

\date{Received: date / Accepted: date}

\maketitle

\begin{abstract}

The combination and aggregation of knowledge from multiple neural networks can be commonly seen in the form of mixtures of experts. However, such combinations are usually done using networks trained on the same tasks, with little mention of the combination of heterogeneous pre-trained networks, especially in the data-free regime. This paper proposes multiple data-free methods for the combination of heterogeneous neural networks, ranging from the utilization of simple output logit statistics, to training specialized gating networks. The gating networks decide whether specific inputs belong to specific networks based on the nature of the expert activations generated. The experiments revealed that the gating networks, including the universal gating approach, constituted the most accurate approach, and therefore represent a pragmatic step towards applications with heterogeneous mixtures of experts in a data-free regime. The code for this project is hosted on github at \url{https://github.com/cwkang1998/network-merging}.

\end{abstract}

\section{Introduction}

The transfer and combination of learning happens often in human learners, allowing the application of existing skills, knowledge and understanding to solve newly encountered problems, commonly under some guidance \citep{perkins1989cognitive}. This process has inspired machine learning methodologies such as transfer learning and multi-task learning, which utilize prior knowledge to improve the learning of novel tasks. However, both
methodologies concern themselves mainly with the improvement of specific tasks \citep{torrey2010transfer,pan2009survey,caruana1997multitask} leveraging additional information acquired from sources that were assumed to have some kind of similarity to their domain, and attempt to exploit this similarity to avoid negative transfer \citep{pan2009survey}. Moreover, in the vast majority of cases, the sharing or combining of knowledge is done during the training process itself, and assumes the availability of sufficient data.

We are motivated by an anticipated future, where the technology landscape is filled with an extremely large number of pre-trained models, many of which might be weakly or negligibly linked to the data sources that were used to train them. This weak linkage could be due to several reasons, no less the fact that many models might have learnt in an unsupervised manner, by freely interacting with an unconstrained environment. So, in a situation where we have several powerful heterogeneous models, but we don't have the data sources that were instrumental to their training, and we can't easily infer how to associate new inputs to specific models (i.e. inputs are thoroughly unknown), how do we then create effective combinations of these models? In other words, we are interested in the problem of combining heterogeneous pre-trained models in a data-free regime, where the inputs are completely unknown. It is important to tackle this problem when the agent designing the mixture of experts is a human being, but it is even more crucial when the {\itshape designer} is an artificial intelligence (AI) agent. Assuming that all ethical considerations and precautions have been duly considered, how should we design an AI agent that is capable of combining heterogeneous pre-trained models in a data-free regime? An AI agent with this capability is useful since it contributes towards artificial general intelligence (AGI), and constitutes a faster approach to building systems for a broader and more diverse set of applications. 

% [Done to a certain degree] Add a paragraph here for the exact scenario: new networks; new input; which network does the input belong to; think of an AI agent that is coordinating this process. AI agent scenario. Could also be faced by humans. Mix human/agent motivation.
%    ### AI agent scenario
% * Table with known/unknown of expert domain (E) and (I) input. We are interested mostly in unknown I, and known/unknown E. The attribution problem for "known I" (what is known is the high-level class, i.e. which expert) cases can be solved (the problem being the automation of the attribution) by training a model based on this knowledge.
% * Both "unknown I and known E" and "unknown I and unknown E" are serious problem when there is a human in the loop, or we are dealing with an AI agent which is trying to combine several pre-trained experts.
% * The above bullet point alludes to the "AI agent scenario". What if we want to design an AI agent with the ability to create other AI agents by recruiting/combining different subsets of experts? When new inputs are received there will no knowledge regarding what experts to use.  

% * [Postponed] Search for iNaturalist's approach. Search for iNaturalist data. How is classification of high-level taxa done? How is the taxonomic hierarchy taken into account?   

Arguably, the two most effective and well-known approaches to combining multiple models, are ensembles \citep{valentini2002ensembles, ju2018relative, minetto2019hydra, shakeel2020automatic} and mixtures of experts \citep{jordan1994hierarchical, nguyen2018practical, wang2018deep, guo-etal-2018-multi, nguyen2019acoustic}. Ensemble learning which is the application of the concept of ``wisdom of the crowd'', is a common method of knowledge consolidation or combination which is widely used \citep{ponti2011combining, sagi2018ensemble}. Through the consolidation of knowledge from multiple learning machines, ensemble learning is able to improve overall performance when compared to the individual members of the ensemble \citep{sagi2018ensemble}. Multiple different types of aggregation methods can be used to consolidate the predictions of the ensemble, each bringing a different benefit to address a certain problem usually related to the bias-variance decomposition \citep{ponti2011combining}, with the most common methods being bagging, boosting and stacking \citep{kittler1998combining, ponti2011combining,sagi2018ensemble}. However, ensemble learning is more commonly used to improve performance on a single task, essentially providing a more robust prediction using the power of the combined learning agents \citep{kittler1998combining, sagi2018ensemble}, each one giving a different interpretation to the same task.

The mixtures of experts (MEs) approach is more in line with our work. A traditional mixture of experts
consists of a set of experts, focusing on equivalent, similar or different tasks, and an additional gating network, which is responsible for deciding which expert(s) should influence the output the most \citep{jordan1994hierarchical}. Both the gating network and all of the experts receive as input the same input pattern. The gating network generally provides a vector of gates, where each gate (a scalar) is multiplied by the output of a corresponding expert, and subsequently all of the modulated outputs are summed, in order to produce the final output. Given the age and flexibility of the approach, there are naturally many variations. Since our aim is to combine pre-trained models, each of which might be responsible for a different task, the ME approach is particularly suitable. In this context, the biggest challenge lies in how to implement and train the gating network, considering that we constrain ourselves to the data-free regime. 

Traditionally the gating network is trained together with the experts, so how do we deal with the situation where the experts have already been trained? In principle one could train a gating network in order to optimize the combination of the pre-trained experts. But, what if the experts are heterogeneous (for different tasks), can be added at any time to the mixture, and we no longer have the original datasets that were used to train the experts? In this case we need the gating network to select a single expert (rather than a combination of experts) for some unknown input. Clearly, training a gating network over this mixture, is no longer feasible. What we need is a gating solution which itself is already pre-trained, allowing  any number of new experts to be added, whereby given a new input, the correct expert that corresponds to that input, can automatically be selected. A solution to this problem, constitutes the main contribution of our paper. In other words, this paper proposes an approach that works towards ``universal gating'' for mixtures of heterogeneous experts, when the experts have already been trained, and when their data is no longer available. For simplicity, and partly because we aim to strengthen the links between Computational Neuroscience and Artificial Intelligence, we restrict our experts to being neural networks. From this point forward, whenever the term ``expert'' is used with respect to our work, the reader can assume that the expert is a neural network.

In our definition, the idea of “universal gating” involves: (1) the possibility of adding any number of new pre-trained heterogeneous experts, (2) one pre-trained gating network for all experts, and (3) the possibility that all experts are trained in a manner that is completely agnostic of any ME architecture or approach. In this paper, we propose two different gating approaches, one which addresses only the first point, whereby each expert has its own gating network, and the other which addresses all three points. For the first approach, we compare two main variants distinguished by how they implement the gating decision, i.e.: (1) via simple logic applied to different statistics over the output logits of the experts, (2) via a gating neural network trained to make a Boolean decision based on expert network activations (including activations of hidden nodes). Since the second variant proved to be significantly more accurate, we used it exclusively in the second approach. The main contributions of this paper include: (1) a novel partially-universal gating approach for mixtures of pre-trained experts (one gating network per expert), (2) a novel universal gating  approach for mixtures of pre-trained experts (one gating network for all experts), and (3) a systematic comparison of different gating variants on multi-task problems constructed from well known datasets.

The next section provides an overview of the work that is most closely related to our proposed approach. The section after that describes the technical details of our approach and our experimental design, and is followed by the results section. In the final sections we discuss our interpretation of the results, and then conclude the paper.

\section{Related Work}

Most previous work on mixtures of experts is concerned with training both the experts and the gating network together for a single task \citep{masoudnia2014mixture}. A smaller subset of research involves using a set of pre-trained experts, and training a gating network to adaptively combine these pre-trained experts again for a particular task \citep{hong2002mixture}. An even smaller subset involves a set of pre-trained experts, a pre-trained gating solution, applicability to heterogeneous tasks, and the possibility of adding any number of new experts \citep{sharma2019Aexpertmatcher, sharma2019Bexpertmatcher}. Our main contribution consists of a novel type of solution to this last scenario, whose properties allow it to be extended to a completely universal gating solution.   

In a traditional gating network, the number of experts is known and therefore the gating network can have a fixed output vector, with each element corresponding to a different expert. This applies even when the number of experts is in the range of thousands \citep{shazeer2017outrageously}. In the latter work, the authors obtain significant efficiency gains by sparsifying the gating outputs, thus requiring only a small set of experts to compute (i.e. those with active gates). Our gating approach does not provide this kind of efficiency (all experts need to compute), however what our approach loses in terms of efficiency, it gains in terms of flexibility, since any number of additional experts can be added any time.

To contrast further with standard mixtures of experts, our gating network does not {\itshape mix experts} (i.e. it does not multiply weighted gates to each expert output vector, and then sum these results). Instead it {\itshape selects one expert}. It does not even select a sparse subset of experts, as with \citep{shazeer2017outrageously}; it only selects one expert. This is related for example to the work of \cite{maeda2020fast} where experts are competitive (i.e. only one expert is chosen to do the final computation). Note also, that as far back as 1991 \citep{jacobs1991adaptive}, a single expert was also selected (in a probabilistic manner). In spite of our current selection of a ``single expert'', our approach can easily be extended to the sparse and mixture cases.

The work by \cite{sharma2019Aexpertmatcher, sharma2019Bexpertmatcher} is closely related to ours, in particular to our partially-universal gating approach, where we use one gating network per expert. In \citep{sharma2019Aexpertmatcher}, each task (or dataset, or expert) is represented by an autoencoder. Each autoencoder can be interpreted as a ``selector'' and is used for selecting the correct model for a new input $x$. The general idea of model selection (termed coarse-level matching in the paper) involves the input $x$ being passed through all of the autoencoders, where the autoencoders with the smaller reconstruction errors are more likely to correspond to the correct models. This same limitation of one ``selector'' per expert can also be seen in \citep{sharma2019Bexpertmatcher}, and is also shared by our partially-universal gating approach. However, in the above work, it is difficult to see how the approach can be extended to the universal case (i.e. one selector for all experts) without fundamentally changing the approach. In our case, the universal generalization is a very simple and natural extension of the basic implementation.

With respect to the nature of the inputs, our approach is also slightly different from standard mixtures of experts. Whereas in the latter case, the gating network will typically receive the same raw input that is fed into the expert networks, in our approach, the inputs of gating networks typically consist of a subset of activations from the expert networks (or features constructed from those activations). In this sense our work is related to \citep{schwab2019granger}, since the gating networks of the latter receive hidden activations as inputs. 

In spite of these localized similarities, to the best of our knowledge there is no other work on gating networks for mixtures of experts, that combines the elements above in order to implement a single universal gating network that allows any number of heterogeneous pre-trained networks to be added at any time, where the heterogeneity can be due to architecture and/or task.

\section{Methods}

\section{Methodology}

\subsection{Datasets, Experts, and Gating}

The experiments utilize the MNIST \citep{mnistDatabase} and CIFAR10 \citep{krizhevsky2009learning} datasets. Two multi-task problems were created using these datasets, namely, the disjoint MNIST problem and the MNIST+CIFAR10 problem. The disjoint MNIST problem focuses on the combination of a network (expert) trained on the first 5 classes of MNIST and a network trained on the last 5 classes of MNIST, which represents the problem of combining heterogeneous networks trained on closely-related tasks. On the other hand, the MNIST+CIFAR10 problem focuses on the combination of a network trained on MNIST and a network trained on CIFAR10, representing the combination of heterogeneous networks trained on distantly related tasks. These problems allow us to compare the effect of task similarity on each combination method, allowing us to better understand the mechanisms of the proposed combination methods.

All of the pre-trained expert networks adopt the LeNet5 convolutional neural network (CNN) architecture \citep{lecun1995comparison}. In general, LeNet5 performs relatively well on MNIST, but the quality of this performance does not extend to the CIFAR10 dataset, for which it exhibits average accuracy. However, as this paper focuses on the performance of the gating solution and its improvement, the architecture is deemed suitable for the experiments. In all cases the networks employ the cross entropy loss for training.

With regards to gating, our general approach entails attributing a new unknown pattern to its corresponding network by observing the node activations of all experts. Our goal is to attribute pattern $p$ to the correct expert $e$, where the {\itshape correct expert} is defined as the network that was trained on data originating from the same distribution that generated $p$. The main assumption underlying our work is that the activations of the {\itshape correct network} are significantly different from the activations of the incorrect networks, and that there are certain unique features regarding {\itshape correct networks} that can be exploited to make the attribution decision. We systematically compare two main approaches which we abbreviate to ``naive concatenation'' and ``smart coordinator'' for the sake of convenience. The former is based on statistics of expert output nodes, and comes in two variants (basic and augmented), whereas the latter consists of training another neural network to associate features of expert activations with a Boolean label which states whether a particular input pattern belongs to an expert or not. The smart coordinator (SC) has two variants, namely SC1 and SC2, where the former uses partially universal gating (i.e. one gating network per expert), whereas SC2 uses a single universal gating network for all experts.

\subsection{Naive concatenation}
\label{sec:naive-cocat}

The simplest way that one can combine multiple heterogeneous networks would be through naive concatenation. Naive concatenation operates with the assumption that the output logits produced by the networks, which refers to the output of the final layer of a network before applying the softmax function, would reflect the ``state of mind'' or uncertainty of the network when given a particular input. In general, if a network is given an input that it {\itshape cannot recognize} it should generally produce logits with values that are closer to each other (i.e. low confidence; high entropy), and if it is given an input that it {\itshape can recognize} it should generally produce logits with one of the values being distinctly higher than the others (i.e. high confidence; low entropy).

To concatenate the heterogeneous expert networks, the involved networks would each be given the same input data to be processed in order to produce the logits. All the produced output logits would then be concatenated together. A statistical function would then be applied on the concatenated logits in order to draw a prediction. Different types of statistical functions were experimented with in order to explore better methods to draw the prediction from the concatenated output logits, namely:

\begin{itemize}
    \item \textbf{Argmax}. The correct network is deemed to be the one with the largest logit.
    \item \textbf{Ratio}. For each expert we take the ratio between each logit and the sum of all logits in that expert. The expert with the largest ratio is deemed the correct expert. 
    \item \textbf{Overall ratio}. The same as the above, except that that each logit is divided by the sum of all logits in the whole concatenation.
    \item \textbf{Third quartile difference}. For each expert, we define this measure as the difference between its max logit, and its third quartile logit. The expert with the largest difference is deemed the correct expert.
    \item \textbf{Standard deviation}. In contradiction to our general intuition pertaining to output entropy and classification confidence, and for comparison purposes, the correct network is deemed to be the one with the smallest standard deviation of logits.
\end{itemize}

\begin{figure}[!htb]
    \centering
    \includegraphics[width=1\textwidth]{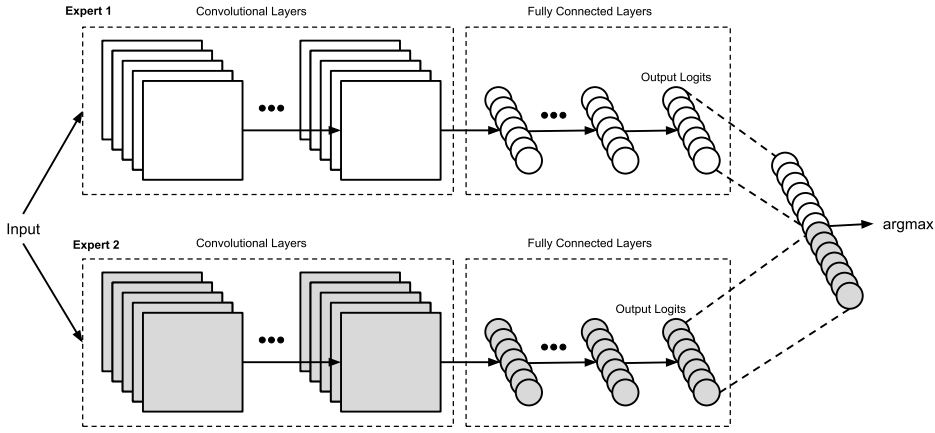}
    \caption{Example naive concatenation.}
\end{figure}

\subsection{Multiple pass with post data augmentation}

One possible issue associated with naive concatenation is the possibility of misclassification due to different experts exhibiting different output distributions. In order to try to counteract this, we experimented with augmenting the input into multiple variants, and then aggregating the results of the multiple decisions obtained by the naive methods described above. The outputs were aggregated either by taking the mean or by a voting mechanism. In  the former case (i.e. mean) we computed the mean of the concatenations (resulting from the multiple augmentations), and then applied our standard decision methods, whereas in the latter case (i.e. voting) we computed decisions on each one of the concatenations, and then voted on the decisions. The augmentation techniques experimented with consisted of sharpening, Gaussian noise, Poisson noise, horizontal flip, vertical flip and random cropping. Note that the data augmentation was not used during expert training, but rather was applied only during the gating decision. 

\begin{algorithm}[H]
    \SetAlgoLined
    \KwData{\\
        Input data: \(x\)\\
        Network 1: \(N_1\)\\
        Network 2: \(N_2\)\\
        Output vector from network: \(O\)\\
        Augmentation functions: \(A \gets [aug_1, aug_2, \cdots, aug_i]\)\\
        Aggregation function: \( agg()\)
    }
    \KwOut{Final prediction, $\lambda$}
    \(Os \gets []\)\; 
    \(O_{N_1}\) $\gets$ \(N_{1}(x)\)\;
    \(O_{N_2}\) $\gets$ \(N_{2}(x)\)\;

    \(Os \gets append(Os, concat(O_{N_1}, O_{N_2}))\);

    \While{not end of list \(A\)}{
    \( x_{aug_i} = A[i](x) \);
    \( O_{N_{1}aug_{i}} \gets N_{1}(x_{aug_i}) \);
    \( O_{N_{2}aug_{i}} \gets N_{2}(x_{aug_i}) \);
    \( Os \gets append(Os, concat(O_{N_{1}aug_{i}}, O_{N_{2}aug_{i}})) \);
    }
    \( O_{combined} \gets agg(Os); \)
    \(\lambda \gets argmax(O_{combined})\);
    \caption{Algorithm for our augmentation approach with naive concatenation.}
\end{algorithm}

Hypothetically, the output vectors of the ``correct network'' should be more consistent, varying minimally across the multiple augmented input passes, while the output vectors of ``incorrect networks'' should differ more across the augmentations, producing different predictions.

\subsection{Smart coordinator}
\label{sec:4.4}

The methods mentioned above are generally concerned with simple statistics of the outputs of the involved heterogeneous networks to draw a combined prediction, which relies greatly on the assumption that the output patterns of the networks behave in a consistent manner. To move beyond this assumption, we propose to train an additional neural network, on features pertaining to key activations of an expert, when observing input pattern $p$, in order to determine whether $p$ is correctly attributed to that expert or not. For the sake of simplicity we term this additional gating network a pattern attribution network (PAN).

\begin{figure}[!htb]
    \centering
    \includegraphics[width=1\textwidth]{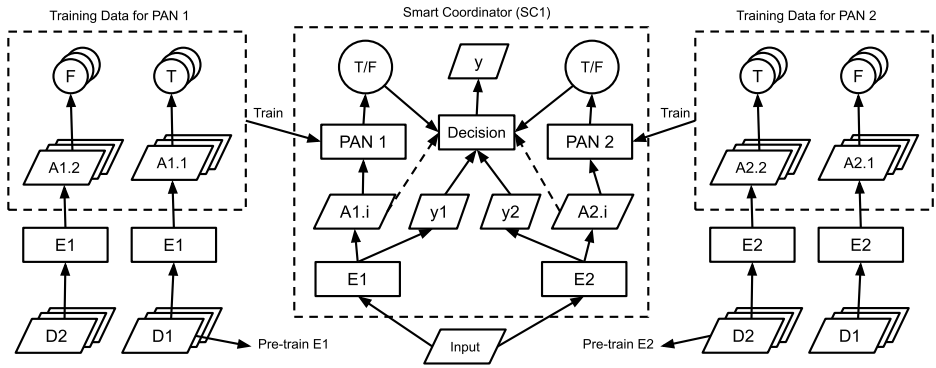}
    \caption{General structure of an SC1 smart coordinator with two PANs. Legend: Di - dataset i; Ei - expert i; Ai.z - activations from expert i after processing dataset z; T/F - true/false; yi - classification result of expert i.}
    \label{fig:SC1}
\end{figure}

For the partially universal gating case (i.e. SC1), each expert has its own PAN, and the combination of all PANs constitutes the ``smart coordinator''. For the sake of clarity, lets assume that we have one expert for classifying cats, and another expert for classifying dogs. Given a new input pattern of a siamese cat, the PAN corresponding to the cat expert should return true, whereas the PAN corresponding to the dog expert should return false. The general architectural structure of an SC1 smart coordinator is shown in Fig. \ref{fig:SC1}.

Each PAN is trained by utilizing the combined datasets of all involved expert networks. Given two heterogeneous networks, network 1 and network 2, the PAN for network 1 (PAN 1) would be trained on the activation features produced by network 1 when given the examples from the combined training dataset of network 1 (i.e. data 1) and network 2 (i.e. data 2). For training PAN 1, the activation features generated from the training dataset of network 1 would be labeled {\itshape true} while the activation features generated from the training dataset of network 2 would be labeled {\itshape false}. A similar training procedure applies to the PAN of network 2 (PAN 2), which would instead have its activation features generated from dataset 1 labeled as {\itshape false} and the activation features generated from dataset 2 labeled as {\itshape true}.

Note that the above account constitutes another non-universal limitation of SC1, since experts are not agnostic of the overall smart coordinator solution, since all expert datasets are involved in the training of each PAN. Note however, that this could partially be addressed if PAN 1 were to be trained with data 1 and other datasets, where the latter are unrelated to the mixture of experts under consideration, and similarly with PAN 2.

Multiple types of activation features were explored in order to find the most effective features that could be used with the PAN. The first and simplest activation features to be tested were the output logits of the network. However, logits alone may not contain enough information for the PAN to recognise a concrete pattern, and therefore we also attempted to explore the outputs from the final fully connected layer. Statistical features of activations (rather than raw activations) were also experimented with in order to gain some insight into the potential usefulness of architecturally agnostic features (contributing further towards universal gating). Examples of the activation statistics include the mean, max and standard deviation. 

With the PANs prepared and trained, the smart coordinator can then be assembled. When given an input, the involved heterogenous networks would process it as usual, however the activation features of each network would be passed to its corresponding PAN to be evaluated. If only one of the PAN returns true, it would imply that the input ``belongs'' to that network, and as such only the output from that network would be taken. However, if both PANs return true, or if both PANs return false, the smart coordinator would default to using the naive concatenation method with the common argmax function for making a combined prediction. \newline

\begin{algorithm}[H]
    \SetAlgoLined
    \KwData{\\
        Input data: \(x\)\\
        Network 1: \(N_1\)\\
        Network 2: \(N_2\)\\
        PAN 1: \(P_1\)\\
        PAN 2: \(P_2\)\\
        Output vector from network: \(O\)\\
        Activation feature from network: \(F\)\\
    }
    \KwOut{Final prediction, $\lambda$}
    \( O_{N_1}, F_{N_1} \gets N_{1}(x) \);
    \( O_{N_2}, F_{N_2} \gets N_{2}(x) \);

    \( O_{P_1} \gets P_{1}(F_{N_1}) \);
    \( O_{P_2} \gets P_{2}(F_{N_2}) \);

    \uIf{\((O_{P_1}\) is \(True\)) and \((O_{P_2}\) is \(False)\)}{
        \(\lambda \gets argmax(O_{N_1})\);
    }\uElseIf{\((O_{P_1}\) is \(False\)) and \((O_{P_2}\) is \(True)\)}{
        \(\lambda \gets argmax(O_{N_2})\);
    }\uElse{
        \(\lambda \gets argmax([ O_{N_1}, O_{N_2}])\);
    }
    \caption{Algorithm for smart coordinator}
\end{algorithm}

\begin{figure}[!htb]
    \centering
    \includegraphics[width=1\textwidth]{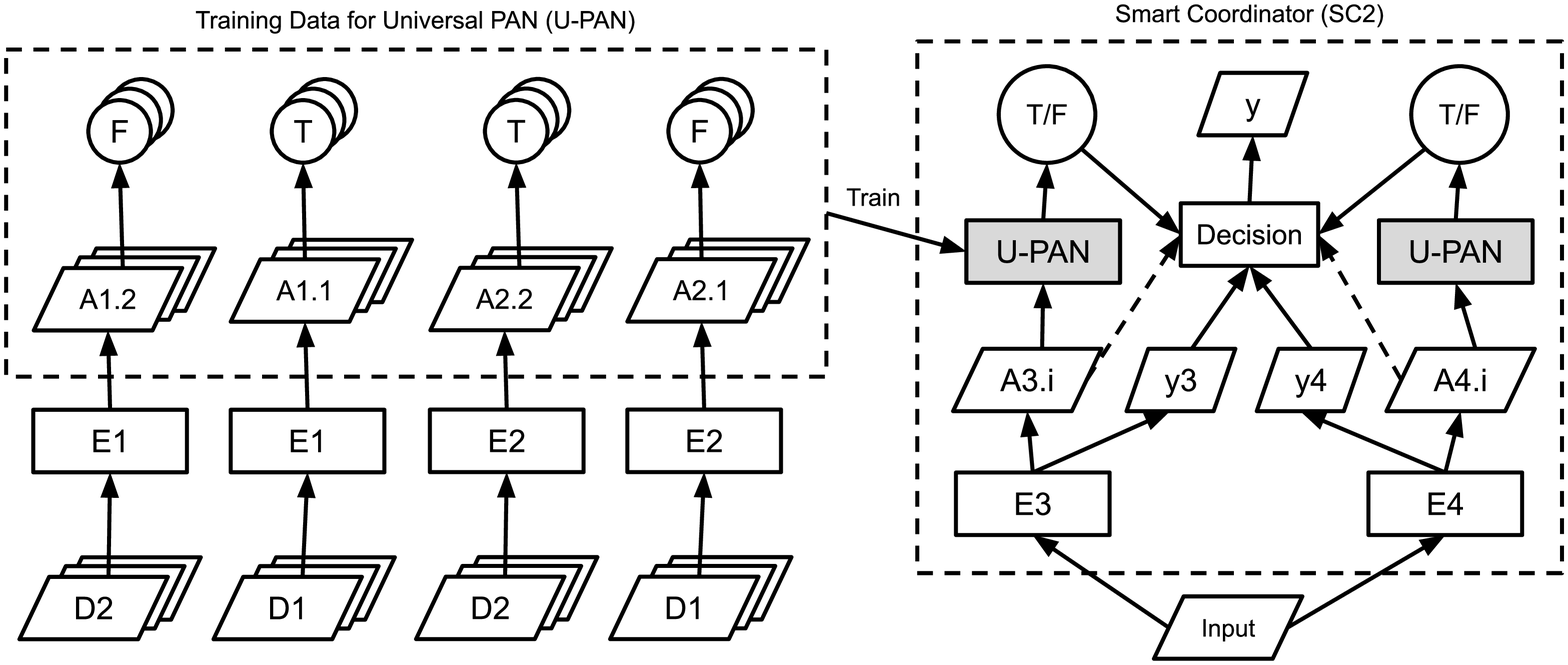}
    \caption{Simplified structure of an SC2 smart coordinator, which by definition involves a single PAN. Legend: Di - dataset i; Ei - expert i; Ai.z - activations from expert i after processing dataset z; T/F - true/false; yi - classification result of expert i.}
    \label{fig:SC2}
\end{figure}

For the fully universal gating solution (i.e. SC2), we take a similar approach, which is depicted in a simplified manner in Fig. \ref{fig:SC2}. The main distinction lies in how we prepare the pattern-attribution dataset, and the fact that we use only one gating network. Briefly, the attribution datasets that are created for individual PANs in SC1 can be combined into one large dataset, with the additional constraint that features of activations (for PAN inputs) need to be abstract enough in order to be applicable to a broad range of new network (expert) architectures. The single dataset is then used to train a single universal PAN (UPAN), which is then combined with experts, their activations, and additional logic for making the final classification decision. The combination of all of these elements constitutes a smart coordinator, in this case SC2. For further universality, the gating network should be trained on activations from experts/datasets completely distinct from those in the mixture of experts under consideration.

The experiments were implemented in Python using the Pytorch framework. All experiments were either executed on
Google Colab, or on the first author's laptop.

\section{Results}

The results were obtained by running experiments across 10 different seeds, each one of which was used for the entire pipeline from training experts to the implementation of the gating solution. In order to better understand the base performance of each expert network after training, we computed test accuracies on their corresponding test sets (i.e. MNIST (0-4), MNIST (5-9), MNIST (0-9), CIFAR10, Fashion-MNIST, and Kuzushiji-MNIST) as depicted in Table \ref{tab:basenet}.

\begin{table}[!htb]
    \caption{Average test accuracy of the expert networks over 10 runs}
    \centering
    \resizebox{0.6\textwidth}{!}{%
        \begin{tabular}{|l|c|}
            \hline
            \textbf{Dataset}         & \textbf{Average Accuracy}  \\ \hline
            First 5 Classes of MNIST & 0.9948                                    \\ \hline
            Last 5 Classes of MNIST  & 0.9879                                    \\ \hline
            MNIST                    & 0.9863                                    \\ \hline
            CIFAR10                  & 0.6305                                    \\ \hline
            Fashion-MNIST            & 0.8708                                    \\ \hline
            Kuzushiji-MNIST          & 0.9180                                    \\ \hline
        \end{tabular}%
    }
    \label{tab:basenet}
\end{table}

Ideally, assuming perfect gating and balanced data, we would want to obtain results with the average accuracy of both experts, that is, the target accuracy can be defined as \(A_t = (A_1 + A_2) / 2\), where \(A_1\) and \(A_2\) refer to the accuracy of the first and second network respectively. The ideal accuracy for the MNIST (0-4) / MNIST (5-9) mixture is thus $0.9914$, and for the MNIST and CIFAR10 mixture is $0.8084$.

When testing the combined experts trained on MNIST and CIFAR10, due to the difference in the number of input channels expected by each expert, with the MNIST expert accepting a single input channel input, and the CIFAR10 expert accepting 3 input channels, the combined testing data from MNIST and CIFAR10 has to be transformed to the correct number of channels for each expert network. More specifically, MNIST images must be transformed from 1 to 3 channels when being fed into the CIFAR10 expert, and CIFAR10 images must be transformed from 3 channels to 1 (i.e. grayscaled) when fed into the MNIST expert. In general, unconstrained mixtures of heterogeneous experts are likely to use many different input representations, and therefore input patterns will frequently need to be transformed and pre-processed with minimal assumptions in order to be compatible with each expert.

Table \ref{tab:logits_stats} summarizes the results of applying different naive concatenation approaches, whereas table \ref{tab:multi_pass} extends these results based on different input augmentations. Table \ref{tab:pan} summarizes the results of individual pattern attribution networks, whereas table \ref{tab:SC1} reports the final smart coordinator results for the partially-universal gating approach (i.e. SC1). In other words, table \ref{tab:pan} depicts how successful PANs are in determining whether specific inputs belong to their experts or not, and table \ref{tab:SC1} depicts the final classification performance after considering (by ``smart coordination'') the PAN attributions and the expert classifications in an overall final decision. Table \ref{tab:upansc2} reports preliminary attribution accuracy results for our universal pattern attribution network together with corresponding smart coordinator accuracies. The column denoted by ``Outputs'' refers to the condition where activation features correspond to output logits, whereas ``Out. (F)'' refers to simple features of the output logits (i.e. mean, max, and standard deviation).

\begin{table}[!htb]
    \caption{Average accuracy of naive concatenation with different statistical functions (argmax, standard deviation, ratio, overall ratio, and third quartile difference)}
    \centering
    \resizebox{\textwidth}{!}{%
        \begin{tabular}{l|l|l|l|l|l|}
            \cline{2-6}
                                                                                                  & \multicolumn{5}{c|}{ \textbf{Statistical Functions}}                                                                           \\ \hline
            \multicolumn{1}{|l|}{\textbf{Problem}} & Argmax                                                                                             & Standard Deviation & Ratio  & Overall Ratio & Third Quartile Difference \\ \hline
            \multicolumn{1}{|l|}{disjoint MNIST (0-4, 5-9)}                                     & 0.9288                                                                                             & 0.0941             & 0.3565 & 0.5085        & 0.9014                    \\ \hline
            \multicolumn{1}{|l|}{MNIST + CIFAR10}                                                 & 0.8039                                                                                             & 0.0026             & 0.2030 & 0.4396        & 0.8061                    \\ \hline
        \end{tabular}%
    }
    \label{tab:logits_stats}
\end{table}

\begin{table}[!htb]
    \caption{Average accuracies of different augmentation approaches with mean and voting aggregations.}
    \centering
    \resizebox{\textwidth}{!}{%
        \begin{tabular}{cl|c|c|}
            \cline{3-4}
            \multicolumn{1}{l}{}                                                                  &                                                                                           & \multicolumn{2}{c|}{\textbf{Aggregation Method}}                   \\ \hline
            \multicolumn{1}{|c|}{\textbf{Problem}} & \multicolumn{1}{c|}{\textbf{Augmentation}} & \textbf{Mean}                                                                                   & \textbf{Voting} \\ \hline
            \multicolumn{1}{|c|}{}                                                                & Single pass sharpen                                                                       & {0.8874}                                                                   & 0.7388          \\ \cline{2-4}
            \multicolumn{1}{|c|}{}                                                                & 5 pass sharpen with differing alpha (0.1, 0.3, 0.5, 0.7, 1.0)                             & 0.4044                                                                                          & 0.1156          \\ \cline{2-4}
            \multicolumn{1}{|c|}{}                                                                & Single pass gaussian noise                                                                & 0.9183                                                                                          & 0.9134          \\ \cline{2-4}
            \multicolumn{1}{|c|}{}                                                                & 5 pass gaussian noise                                                                     & 0.9119                                                                                          & 0.9108          \\ \cline{2-4}
            \multicolumn{1}{|c|}{}                                                                & \begin{tabular}[c]{@{}l@{}}6 pass gaussian noise with differing standard deviation \\ (0.05, 0.1, 0.3, 0.5, 0.7, 1)\end{tabular}                                                                & 0.9261                                                                                          & \textbf{0.9280}          \\ \cline{2-4}
            \multicolumn{1}{|c|}{}                                                                & Single pass poisson noise                                                                 & 0.9193                                                                                          & 0.9194          \\ \cline{2-4}
            \multicolumn{1}{|c|}{}                                                                & 5 pass poisson noise                                                                      & 0.9101                                                                                          & 0.9085          \\ \cline{2-4}
            \multicolumn{1}{|c|}{}                                                                & \begin{tabular}[c]{@{}l@{}}6 pass poisson noise with differing standard deviation \\ (0.05, 0.1, 0.3, 0.5, 0.7, 1)\end{tabular}                                                                & 0.9156                                                                                          & 0.9182          \\ \cline{2-4}
            \multicolumn{1}{|c|}{}                                                                & Horizontal and Vertical flip                                                              & 0.7898                                                                                          & 0.1045          \\ \cline{2-4}
            \multicolumn{1}{|c|}{{MNIST (0-4/5-9)}}                  & Random cropping                                                                           & 0.8957                                                                                          & 0.7289          \\ \hline
            \multicolumn{1}{|c|}{}                                                                & Single pass sharpen                                                                       & 0.7383                                                                                          & 0.5915          \\ \cline{2-4}
            \multicolumn{1}{|c|}{}                                                                & 5 pass sharpen with differing alpha (0.1, 0.3, 0.5, 0.7, 1.0)                             & 0.4320                                                                                          & 0.0942          \\ \cline{2-4}
            \multicolumn{1}{|c|}{}                                                                & Single pass gaussian noise                                                                & 0.7467                                                                                          & 0.7209          \\ \cline{2-4}
            \multicolumn{1}{|c|}{}                                                                & 5 pass gaussian noise                                                                     & 0.6945                                                                                          & 0.6805          \\ \cline{2-4}
            \multicolumn{1}{|c|}{}                                                                & \begin{tabular}[c]{@{}l@{}}6 pass gaussian noise with differing standard deviation \\ (0.05, 0.1, 0.3, 0.5, 0.7, 1)\end{tabular}                                                                & 0.7794                                                                                          & \textbf{0.7951}          \\ \cline{2-4}
            \multicolumn{1}{|c|}{}                                                                & Single pass poisson noise                                                                 & 0.7845                                                                                          & 0.7616          \\ \cline{2-4}
            \multicolumn{1}{|c|}{}                                                                & 5 pass poisson noise                                                                      & 0.7663                                                                                          & 0.7583          \\ \cline{2-4}
            \multicolumn{1}{|c|}{}                                                                & \begin{tabular}[c]{@{}l@{}}6 pass poisson noise with differing standard deviation\\ (0.05, 0.1, 0.3, 0.5, 0.7, 1)\end{tabular}                                                                & 0.7781                                                                                          & 0.7777          \\ \cline{2-4}
            \multicolumn{1}{|c|}{}                                                                & Horizontal and Vertical flip                                                              & 0.6765                                                                                          & 0.2654          \\ \cline{2-4}
            \multicolumn{1}{|c|}{{MNIST/CIFAR10}}                              & Random cropping                                                                           & 0.7538                                                                                          & 0.5295          \\ \hline
        \end{tabular}%
    }
    \label{tab:multi_pass}
\end{table}

\begin{table}[!htb]
    \caption{Average pattern attribution accuracy for PAN}
    \centering
    \resizebox{\textwidth}{!}{%
        \begin{tabular}{lll|c|c|c|}
            \cline{4-6}
                                                                                                           &                                                                                               &                                                                                              & \multicolumn{3}{c|}{\textbf{Coordinator feature}}                                                              \\ \hline
            \multicolumn{1}{|c|}{\textbf{Positive Dataset}} & \multicolumn{1}{c|}{\textbf{Negative Dataset}} & \multicolumn{1}{c|}{\textbf{Testing Dataset}} & \textbf{Outputs}                                                                    & \textbf{Final FC} & \textbf{Out. (F)} \\ \hline
            \multicolumn{1}{|l|}{MNIST (0-4)}                                            & \multicolumn{1}{l|}{MNIST (5-9)}                                             & MNIST                                                                                        & {0.9254}                                                                    & \textbf{0.9761}               & 0.8655                              \\ \hline
            \multicolumn{1}{|l|}{MNIST (5-9)}                                             & \multicolumn{1}{l|}{MNIST (0-4)}                                            & MNIST                                                                                        & 0.8943                                                                                           & \textbf{0.9755}               & 0.8419                              \\ \hline
            \multicolumn{1}{|l|}{MNIST}                                                                    & \multicolumn{1}{l|}{CIFAR10}                                                                  & MNIST/CIFAR10                                                                              & \textbf{0.9999}                                                                                           & \textbf{0.9999}               & \textbf{0.9999}                              \\ \hline
            \multicolumn{1}{|l|}{CIFAR10}                                                                  & \multicolumn{1}{l|}{MNIST}                                                               & MNIST/CIFAR10                                                                              & 0.9983                                                                                           & \textbf{0.9998}               & 0.9716                              \\ \hline
        \end{tabular}%
    }
    \label{tab:pan}
\end{table}

\begin{table}[!htb]
    \caption{Average smart coordination accuracy for SC1}
    \centering
    \resizebox{0.7\textwidth}{!}{%
        \begin{tabular}{c|c|c|l}
            \cline{2-4}
            \multicolumn{1}{l|}{}                                                                  & \multicolumn{3}{c|}{ \textbf{Coordinator feature}}                                                                                   \\ \hline
            \multicolumn{1}{|c|}{\textbf{Problems}} & \textbf{Outputs}                                                                    & \textbf{Final FC} & \multicolumn{1}{c|}{\textbf{Out. (F)}} \\ \hline
            \multicolumn{1}{|c|}{disjoint MNIST (0-4, 5-9)}                                      & {0.9458}                                                                    & \textbf{0.9733}               & \multicolumn{1}{l|}{0.9294}                              \\ \hline
            \multicolumn{1}{|c|}{MNIST + CIFAR10}                                                  & 0.8083                                                                                           & \textbf{0.8084}               & \multicolumn{1}{l|}{0.8071}                              \\ \hline
        \end{tabular}%
    }
    \label{tab:SC1}
\end{table}

\begin{table}[!htb]
    \caption{UPAN and SC2 performance. Results are portrayed as $x/y$ where x denotes the average pattern attribution accuracy for UPAN, and y denotes the corresponding smart coordinator (SC2) accuracy. ``Fashion'' denotes the Fashion-MNIST dataset, whereas ``Kuzushiji'' refers to the Kuzushiji-MNIST dataset.}
    \centering
    \resizebox{0.85\textwidth}{!}{%
        \begin{tabular}{c|c|c|l}
            \cline{3-4}
            \multicolumn{2}{l|}{}                                                                  & \multicolumn{2}{c|}{ \textbf{Activation feature}}                                                                                   \\ \hline
            \multicolumn{1}{|c|}{\textbf{PAN trained on}} & \textbf{PAN tested on} & \textbf{Outputs}                       &  \multicolumn{1}{c|}{\textbf{Out. (F)}} \\ \hline
            \multicolumn{1}{|c|}{MNIST (0-4, 5-9)} & {MNIST (0-4, 5-9)} & {\textbf{0.9566 / 0.9515}} & \multicolumn{1}{c|}{0.9386 / 0.9363}                              \\ \hline
            \multicolumn{1}{|c|}{MNIST + CIFAR10} & {MNIST + CIFAR10} & {\textbf{0.9998 / 0.8874}}                                            & \multicolumn{1}{c|}{0.9932 / 0.8844} \\ \hline
            \multicolumn{1}{|c|}{MNIST (0-4, 5-9)} & {MNIST+CIFAR10} & {N/A}                                                                    & \multicolumn{1}{c|}{\textbf{0.8859 / 0.8762}}  \\ \hline
            \multicolumn{1}{|c|}{MNIST+CIFAR10} & {MNIST(0-4,5-9)} & {N/A}                                                                    & \multicolumn{1}{c|}{\textbf{0.8238 / 0.8155}}  \\ \hline
            \multicolumn{1}{|c|}{MNIST(0-4, 5-9)} & {Fashion+Kuzushiji} & {N/A}                                          & \multicolumn{1}{c|}{\textbf{0.8763 / 0.8177}} \\ \hline
            \multicolumn{1}{|c|}{MNIST+CIFAR10} & {Fashion+Kuzushiji} & {0.8103 / 0.7515}                                          & \multicolumn{1}{c|}{\textbf{0.8797 / 0.8204}} \\ \hline
        \end{tabular}%
    }
    \label{tab:upansc2}
\end{table}

% \begin{table}[!htb]
%     \caption{Average smart coordination accuracy (SC2).}
%     \centering
%     \resizebox{0.7\textwidth}{!}{%
%         \begin{tabular}{c|c|c|l}
%             \cline{2-4}
%             \multicolumn{1}{l|}{}                                                                  & \multicolumn{3}{c|}{ \textbf{Coordinator feature}}                                                                                   \\ \hline
%             \multicolumn{1}{|c|}{\textbf{Problems}} & \textbf{Outputs}                                                                    & \textbf{Final FC} & \multicolumn{1}{c|}{\textbf{Out. (F)}} \\ \hline
%             \multicolumn{1}{|c|}{disjoint MNIST (0-4, 5-9)}                                      & {x}                                                                    & \textbf{x}               & \multicolumn{1}{l|}{x}                              \\ \hline
%             \multicolumn{1}{|c|}{MNIST + CIFAR10}                                                  & x                                                                                           & \textbf{x}               & \multicolumn{1}{l|}{x}                              \\ \hline
%         \end{tabular}%
%     }
%     \label{tab:SC2}
% \end{table}

\section{Discussion}

The naive concatenation methods were effectively the main control conditions for our experiments. The argmax function in particular, was initially expected to perform poorly, especially on similar tasks such as the disjoint MNIST problem, due to it simply taking the argmax of the concatenated outputs of multiple networks. However, it actually performed the best amongst all the statistical functions, and outperformed all of the multi-pass augmentation approaches as well. When comparing argmax with the ideal target combination accuracy (estimated by averaging individual accuracies), the accuracy of the naive concatenation method with the argmax statistical function was lower, as expected, but the difference between them was not very large, with the disjoint MNIST problem having a 0.063 difference, and the MNIST + CIFAR10 problem only having a 0.0045 difference due to it being the combination of two less similar tasks.

Another interesting observation can be seen in naive concatenation with third quartile difference, which performed worse than argmax on the disjoint MNIST problem, but when applied to the MNIST + CIFAR10 problem, it  actually performed better than argmax, although it still did not reach the ideal target combination accuracy, lacking by 0.002. Judging by these observations, the assumption that the logits of a network reflect the ``state of mind'' of a network does partially hold true, with distinctly large activations more likely to correspond to correct predictions and thus indicating the ``true'' networks. However, looking at the performance of other naive mechanisms, namely standard deviation, ratio, and overall ratio, we can tentatively conclude that their underlying hypothesis that the remaining logits (i.e. not max) also contain useful information for the  gating decision is not entirely confirmed, and that more work needs to be done if this information is to be extracted and used in a simplistic manner.

The multiple pass approach with augmentation is an extension to the naive concatenation with argmax method, taking its inspiration from ensemble learning, but instead of passing the input to multiple trained networks with individual strength and then consolidating their strengths to make a single prediction on the same task, it provides the same network with multiple views or perspectives on the same input data, allowing multiple predictions or ``opinions'' to be formed. The augmentation is meant to provide different perspectives however it was found it does not actually improve the predictions when aggregated. In fact, all of the augmentations caused a drop in accuracy, across both aggregation methods. The approach with 6-pass gaussian noise with differing standard deviations performs with an accuracy closest to the argmax baseline, however this might simply be due to the passes having less gaussian noise with the lower standard deviation, thus causing the augmented image to look similar to the original input after augmentation, which tips the majority vote to the predicted classes of the original data, causing it to resemble the naive concatenation baseline. As such, it can be concluded that the use of post data augmentation here does not actually highlight certain features out for the network to provide additional ``opinions'' or information that can contribute to a better rounded prediction, but instead simply confuses the network causing it to perform less accurately and have a less confident prediction in the end. Note that the base networks used for this experiment were not trained with data augmentation, and as such if the networks were trained using some sort of data augmentation, the result of this experiment could potentially turn out differently.

From observing the test accuracy obtained from the smart coordinator SC1, it is apparent that the individual PANs were able to successfully learn the activation patterns from the individual networks, in order to identify the attribution of a given input to its corresponding network. When trained on the crafted feature dataset, the performance of PAN using logits and hidden features generally achieved accuracy greater than 0.9, with only the PAN for the last 5 MNIST class using logits as features having a lower accuracy at 0.8943. For PANs using logit activation statistics instead of raw features from the network, the accuracy on the crafted dataset for the disjoint MNIST problem when tested was significantly lower than its counterpart, having accuracy of 0.8655 and 0.8415 for the first 5 and last 5 MNIST classes respectively. Similar to the naive concatenation baseline, SC1 seems to work better in coordinating distinct tasks compared to similar tasks, with accuracy close to 100\% when trained on the crafted dataset for the MNIST+CIFAR10 problem. Determining the attribution of a given input seems to be harder when the experts involved are trained on similar data, as the corresponding PANs have to learn specific features corresponding to the smaller differences between the different data distributions.

All types of PANs in the SC1 smart coordinator resulted in an overall improvement in the accuracy for all problems when compared to naive concatenation using argmax. The output logit based PAN showed a performance increase of 0.017 for the disjoint MNIST problem and a performance increase of 0.0044 for the MNIST+CIFAR10 problem. The final FC based PAN performed the best, with a performance increase of 0.0445 for the disjoint MNIST problem and a performance increase of 0.0045 for the MNIST+CIFAR10 problem. The PAN based on output logit statistics exhibited the lowest increase in performance, having only a performance increase of 0.0006 for the disjoint MNIST problem and performance increase of 0.0032 for the MNIST+CIFAR10 problem. The good performance of the final FC based PAN is most probably due to the availability of more information as compared to the other types of PAN, and similarly, the relatively poor performance of the output logit statistics is most probably due to a lack of discriminatory information. One final observation taken from the experiment on the final FC based PAN is that it reached the ideal target combination accuracy for the MNIST+CIFAR10 problem.

Preliminary results on universal pattern attribution networks (UPANs) depicted in Table \ref{tab:upansc2} are indicative of the usefulness of the approach. Rows 3-6 exemplify cases that include, or are entirely made of, never before seen datasets, and therefore serve as the ultimate validation of the approach. Although attribution accuracy is lower relative to the easier conditions depicted in rows 1-2, the UPAN is still capable of performing significantly above chance level. The cells depicted by {\itshape N/A} refer to cases where the structure of output vectors are inconsistent between training and test cases (e.g. 5 outputs vs. 10 outputs), and therefore serve as a reminder of the importance of architecturally agnostic features. Additional unreported experiments were also conducted on what we call fast PANs (FPANs) whereby we use a UPAN to train a separate network to associate input images directly with specific experts, based on UPAN decisions. Preliminary results were positive, even outperforming the original UPAN accuracy. This development is important, because apart from its implications for accuracy, it represents a significant improvement in efficiency, given that we no longer need inputs to be processed by each expert in order for a gating decision to be made.

\section{Conclusion}

This paper has explored multiple approaches for the combination of heterogeneous pre-trained neural networks in a data-free regime, culminating in a universal gating network. Through the reported experiments, it was shown that it is possible to use neural activations themselves, or abstract features of these, in order to infer whether a particular input pattern belongs to a network/expert or not.   

Given the potential of the universal gating approach to the rapid development of complex applications based on mixtures of heterogeneous experts, we recommend further research into: (1) training and testing on an even broader range of datasets and neural architectures, (2) more diverse and discriminatory network-agnostic features, and (3) fast pattern attribution networks (FPANs) where UPANs are used to train an additional network to associate inputs directly with gating decisions thus avoiding the need for inputs to be processed by all experts before making an attribution decision.

\section*{Funding}
Not applicable.

\section*{Conflicts of interest/Competing interests}
Not applicable.

\section*{Availability of data and material}
All datasets used are in the public domain.

\section*{Code availability}
https://github.com/cwkang1998/network-merging


\bibliography{univgatingbib} 

\begin{thebibliography}{10}

\bibitem{perkins1989cognitive}
D.~N. Perkins and G.~Salomon, ``Are cognitive skills context-bound?,'' {\em
  Educational researcher}, vol.~18, no.~1, pp.~16--25, 1989.

\bibitem{torrey2010transfer}
L.~Torrey and J.~Shavlik, ``Transfer learning,'' in {\em Handbook of research
  on machine learning applications and trends: algorithms, methods, and
  techniques}, pp.~242--264, IGI Global, 2010.

\bibitem{pan2009survey}
S.~J. Pan and Q.~Yang, ``A survey on transfer learning,'' {\em IEEE
  Transactions on knowledge and data engineering}, vol.~22, no.~10,
  pp.~1345--1359, 2009.

\bibitem{caruana1997multitask}
R.~Caruana, ``Multitask learning,'' {\em Machine learning}, vol.~28, no.~1,
  pp.~41--75, 1997.

\bibitem{valentini2002ensembles}
G.~Valentini and F.~Masulli, ``Ensembles of learning machines,'' in {\em
  Italian workshop on neural nets}, pp.~3--20, Springer, 2002.

\bibitem{ju2018relative}
C.~Ju, A.~Bibaut, and M.~van~der Laan, ``The relative performance of ensemble
  methods with deep convolutional neural networks for image classification,''
  {\em Journal of Applied Statistics}, vol.~45, no.~15, pp.~2800--2818, 2018.

\bibitem{minetto2019hydra}
R.~Minetto, M.~P. Segundo, and S.~Sarkar, ``Hydra: An ensemble of convolutional
  neural networks for geospatial land classification,'' {\em IEEE Transactions
  on Geoscience and Remote Sensing}, vol.~57, no.~9, pp.~6530--6541, 2019.

\bibitem{shakeel2020automatic}
P.~M. Shakeel, A.~Tolba, Z.~Al-Makhadmeh, and M.~M. Jaber, ``Automatic
  detection of lung cancer from biomedical data set using discrete adaboost
  optimized ensemble learning generalized neural networks,'' {\em Neural
  Computing and Applications}, vol.~32, no.~3, pp.~777--790, 2020.

\bibitem{jordan1994hierarchical}
M.~I. Jordan and R.~A. Jacobs, ``Hierarchical mixtures of experts and the em
  algorithm,'' {\em Neural computation}, vol.~6, no.~2, pp.~181--214, 1994.

\bibitem{nguyen2018practical}
H.~D. Nguyen and F.~Chamroukhi, ``Practical and theoretical aspects of
  mixture-of-experts modeling: An overview,'' {\em Wiley Interdisciplinary
  Reviews: Data Mining and Knowledge Discovery}, vol.~8, no.~4, p.~e1246, 2018.

\bibitem{wang2018deep}
X.~Wang, F.~Yu, L.~Dunlap, Y.-A. Ma, R.~Wang, A.~Mirhoseini, T.~Darrell, and
  J.~E. Gonzalez, ``Deep mixture of experts via shallow embedding,'' {\em arXiv
  preprint arXiv:1806.01531}, 2018.

\bibitem{guo-etal-2018-multi}
J.~Guo, D.~Shah, and R.~Barzilay, ``Multi-source domain adaptation with mixture
  of experts,'' in {\em Proceedings of the 2018 Conference on Empirical Methods
  in Natural Language Processing}, (Brussels, Belgium), pp.~4694--4703,
  Association for Computational Linguistics, Oct.-Nov. 2018.

\bibitem{nguyen2019acoustic}
T.~Nguyen and F.~Pernkopf, ``Acoustic scene classification with mismatched
  recording devices using mixture of experts layer,'' in {\em 2019 IEEE
  International Conference on Multimedia and Expo (ICME)}, pp.~1666--1671,
  IEEE, 2019.

\bibitem{ponti2011combining}
M.~P. Ponti~Jr, ``Combining classifiers: from the creation of ensembles to the
  decision fusion,'' in {\em 2011 24th SIBGRAPI Conference on Graphics,
  Patterns, and Images Tutorials}, pp.~1--10, IEEE, 2011.

\bibitem{sagi2018ensemble}
O.~Sagi and L.~Rokach, ``Ensemble learning: A survey,'' {\em Wiley
  Interdisciplinary Reviews: Data Mining and Knowledge Discovery}, vol.~8,
  no.~4, p.~e1249, 2018.

\bibitem{kittler1998combining}
J.~Kittler, M.~Hatef, R.~P. Duin, and J.~Matas, ``On combining classifiers,''
  {\em IEEE transactions on pattern analysis and machine intelligence},
  vol.~20, no.~3, pp.~226--239, 1998.

\bibitem{masoudnia2014mixture}
S.~Masoudnia and R.~Ebrahimpour, ``Mixture of experts: a literature survey,''
  {\em Artificial Intelligence Review}, vol.~42, no.~2, pp.~275--293, 2014.

\bibitem{hong2002mixture}
X.~Hong and C.~J. Harris, ``A mixture of experts network structure construction
  algorithm for modelling and control,'' {\em Applied intelligence}, vol.~16,
  no.~1, pp.~59--69, 2002.

\bibitem{sharma2019Aexpertmatcher}
V.~Sharma, P.~Vepakomma, T.~Swedish, K.~Chang, J.~Kalpathy-Cramer, and
  R.~Raskar, ``Expertmatcher: Automating ml model selection for users in
  resource constrained countries,'' {\em arXiv preprint arXiv:1910.02312},
  2019.

\bibitem{sharma2019Bexpertmatcher}
V.~Sharma, P.~Vepakomma, T.~Swedish, K.~Chang, J.~Kalpathy-Cramer, and
  R.~Raskar, ``Expertmatcher: Automating ml model selection for clients using
  hidden representations,'' {\em arXiv preprint arXiv:1910.03731}, 2019.

\bibitem{shazeer2017outrageously}
N.~Shazeer, A.~Mirhoseini, K.~Maziarz, A.~Davis, Q.~Le, G.~Hinton, and J.~Dean,
  ``Outrageously large neural networks: The sparsely-gated mixture-of-experts
  layer,'' {\em arXiv preprint arXiv:1701.06538}, 2017.

\bibitem{maeda2020fast}
S.~Maeda, ``Fast and flexible image blind denoising via competition of
  experts,'' in {\em Proceedings of the IEEE/CVF Conference on Computer Vision
  and Pattern Recognition Workshops}, pp.~528--529, 2020.

\bibitem{jacobs1991adaptive}
R.~A. Jacobs, M.~I. Jordan, S.~J. Nowlan, and G.~E. Hinton, ``Adaptive mixtures
  of local experts,'' {\em Neural computation}, vol.~3, no.~1, pp.~79--87,
  1991.

\bibitem{schwab2019granger}
P.~Schwab, D.~Miladinovic, and W.~Karlen, ``Granger-causal attentive mixtures
  of experts: Learning important features with neural networks,'' in {\em
  Proceedings of the AAAI Conference on Artificial Intelligence}, vol.~33,
  pp.~4846--4853, 2019.

\bibitem{mnistDatabase}
C.~C. Yann~Lecun and C.~Burges, ``The mnist database.''

\bibitem{krizhevsky2009learning}
A.~Krizhevsky, G.~Hinton, {\em et~al.}, ``Learning multiple layers of features
  from tiny images,'' 2009.

\bibitem{lecun1995comparison}
Y.~LeCun, L.~Jackel, L.~Bottou, A.~Brunot, C.~Cortes, J.~Denker, H.~Drucker,
  I.~Guyon, U.~Muller, E.~Sackinger, {\em et~al.}, ``Comparison of learning
  algorithms for handwritten digit recognition,'' in {\em International
  conference on artificial neural networks}, vol.~60, pp.~53--60, Perth,
  Australia, 1995.

\end{thebibliography}
\bibliographystyle{ieeetr}
% \bibliographystyle{agsm}
%\bibliographystyle{ksfh_nat}





% LNCS Homepage, \url{http://www.springer.com/lncs}. Last accessed 4
% \end{thebibliography}

\end{document}